\documentclass[runningheads]{llncs}
\usepackage[T1]{fontenc}
\usepackage{graphicx}
\usepackage{subcaption}
\usepackage{comment}
\usepackage{amsmath}
\usepackage{booktabs}
\usepackage{caption}
\usepackage{wrapfig}
\usepackage{grffile}
\usepackage{hyperref}

\begin{document}
\title{A Novel Approach to Breast Cancer Segmentation using U-Net Model with Attention Mechanisms and FedProx}

\author{Eyad Gad\inst{1}\and Mustafa Abou Khatwa\inst{1}\and Mustafa A. Elattar\inst{2,3}\and Sahar Selim\inst{2,3}}

\authorrunning{E. Gad et al.}
\titlerunning{A Novel Approach to Breast Cancer Segmentation}
\institute{School of Engineering and Applied Sciences, Nile University, Egypt\and
Medical Imaging and Image Processing Research Group, Center for Informatics Science, Nile University, Egypt\and
School of Information Technology and Computer Science, Nile University, Egypt\\
\email{\{e.gad, m.aboukatwa, melattar, sselim\}@nu.edu.eg}}
\maketitle               
\begin{abstract}
Breast cancer is a leading cause of death among women worldwide, emphasizing the need for early detection and accurate diagnosis. As such Ultrasound Imaging, a reliable and cost-effective tool, is used for this purpose, however the sensitive nature of medical data makes it challenging to develop accurate and private artificial intelligence models. A solution is Federated Learning as it is a promising technique for distributed machine learning on sensitive medical data while preserving patient privacy. However, training on non-Independent and non-Identically Distributed (non-IID) local datasets can impact the accuracy and generalization of the trained model, which is crucial for accurate tumour boundary delineation in BC segmentation. This study aims to tackle this challenge by applying the Federated Proximal (FedProx) method to non-IID Ultrasonic Breast Cancer Imaging datasets. Moreover, we focus on enhancing tumour segmentation accuracy by incorporating a modified U-Net model with attention mechanisms. Our approach resulted in a global model with 96\% accuracy, demonstrating the effectiveness of our method in enhancing tumour segmentation accuracy while preserving patient privacy. Our findings suggest that FedProx has the potential to be a promising approach for training precise machine learning models on non-IID local medical datasets.

\end{abstract}
\section{Introduction}

Breast Cancer (BC) is a major public health concern worldwide where it affects millions of people, and early detection is crucial for successful treatment while increasing the likelihood of a patient's survival. According to The National Breast Cancer Foundation, and The Center for Disease Control and Prevention (CDC). BC is the most common cancer among women in the USA, with a rate of 129.7 new cases per 100,000 in 2019 while being the second leading cause of cancer related deaths among women, at a rate of 19.4 per 100,000 in 2019. Globally, BC is the most common cancer diagnosed among women, with an estimated 2.3 million new cases diagnosed in 2020.  Preventative measures and early detection strategies, such as regular screening, can help reduce the impact of BC on people\textquotesingle s lives, according to the World Health Organization (WHO) \cite{Ref_1,Ref_2,Ref_3}. Hence, the early detection and accurate diagnosis of BC is a critical affair for effective treatment and improved patient health. Modern diagnostic approaches utilize Ultrasonic Images (USI), where Ultrasonic (US) sensors send acoustic waves into the subject to determine distance, composition, and density. US sensors have become valuable tools for BC screening, as they can detect lesions that may not be visible on mammography, which is a radiological examination of breasts used to detect cancers \cite{Moghbel_Ooi_Ismail_Hau_Memari_2019}. 

Clinics and hospitals often face difficulties in accurately interpreting and diagnosing breast cancer from medical images. To address this challenge, Artificial Intelligence (AI) has been employed to enhance diagnosis and classification processes . Deep Learning (DL) algorithms have proven effective in extracting valuable information from medical images, enabling tasks such as segmenting and classifying Ultrasound Breast Cancer Images (USBCI). These AI-driven approaches have significantly improved the accuracy of diagnosis and treatment planning, as highlighted in a study by Xu et al. Additionally, several research studies have demonstrated that AI-based models for lesion detection and classification can enhance the sensitivity and specificity of breast cancer diagnosis, leading to improved outcomes \cite{XU20191,Vakanski_Xian_Freer_2020,LYU2023104425,9676574,FedZaCt2022,FedMix:10004567,Mouhni_Elkalay_Chakraoui_Abdali_Ammoumou_Amalou_2022,BreastCancerClassification:2022}.

The incorporation of AI has demonstrated promising potential in augmenting the capabilities of healthcare professionals in the field of BC diagnosis. However, as imaging technology advances and the volume and quality of imaging data increase, DL algorithms like ResNet, U-Net, and V-Net face challenges in accurately segmenting USBCI. Although the U-Net model is widely recognized for its effectiveness in cancer segmentation, it encounters difficulties in capturing intricate details and handling complex image structures. To overcome these limitations, this paper proposes the adoption of the Attention U-Net model. By incorporating attention mechanisms into the U-Net architecture, the model gains the ability to selectively focus on important regions within the image while disregarding irrelevant information. This attention-based approach enhances the U-Net's capability to capture fine-grained details, resulting in more precise and accurate segmentation outcomes, however, effectively analyzing such extensive datasets can be time-consuming and resource-intensive. Additionally, concerns regarding privacy arise when dealing with large amounts of data stored in a centralized location\cite{XU20191,Vakanski_Xian_Freer_2020,9676574,lazo2020comparison}.

To address these challenges, FL is proposed, allowing the training of machine learning models on distributed datasets across different locations without sharing the raw data. In the medical field, FL is employed to maintain the privacy of medical data and alleviate network strain by training a predictor in a distributed manner, rather than transmitting raw data to a central server. This setup involves remote devices periodically communicating with a central server to learn a global model. In each communication round, a subset of selected edge devices conducts local training using their non-identically distributed user data and sends local updates to the server. Upon receiving the updates, the server incorporates them and returns the updated global model to another subset of devices. This iterative training process continues throughout the network until convergence, or a stopping criterion is met. This approach preserves privacy, minimizes resource consumption, and enables efficient utilization of local resources. Additionally, many researchers have studied the potential of FL for medical image analysis, including segmentation, classification, and synthesis \cite{9676574,yang2022robust,ANovelMultistageTransfer:2022}.

However, FL, with its involvement of multiple parties and the non-independent and non-identically distributed (non-IID) nature of data generated and collected across the network, presents specific challenges, particularly in medical image tasks such as diagnosis and segmentation, including those related to USBCI. These challenges are further compounded by the significant variations in the data characteristics used for training, which stem from different imaging devices. Consequently, the datasets utilized in FL exhibit diverse properties, including variations in quality, resolution, and contrast. This data generation paradigm violates commonly used assumptions of independent and identically distributed (IID) data in distributed optimization introduces complexities, such as an increased likelihood of encountering stragglers and added intricacies in terms of modeling, analysis, and evaluation.
Hence, to address the challenges of developing accurate and private AI models for BC segmentation using non-IID USBCI datasets, his study proposes the utilization of FL with the FedProx method. \cite{RothHolger10.1007/978-3-030-60548-3_18,li2020federated}. FedProx has the potential to be a promising approach for training precise models on non-IID medical datasets. The proposed approach is implemented in a setup involving three nodes or clients, leveraging the availability of datasets from two distinct sources, which were introduced in \cite{Yap_Pons_Marti_Ganau_Sentis_Zwiggelaar_Davison_Marti_2018,Al-Dhabyani_Gomaa_Khaled_Fahmy_2020}.  Furthermore, data augmentation techniques are applied to enhance the datasets, contributing to the overall improvement of the BC segmentation model's performance.

The main aim of the paper lies in improving the accuracy of breast tumor segmentation. To achieve this, the paper proposes an enhanced approach that involves modifying the U-Net model with attention mechanisms. This attention U-Net model serves as a segmentation DL model utilized by three clients within the FL framework. Additionally, the paper employs the FedProx method to aggregate the trained weights of the models from the clients, ensuring collaborative learning and integration of knowledge from the distributed clients while preserving privacy and security.

\section{Literature Review}

Over the past few years, several FL techniques have been proposed to enhance the accuracy and efficiency of medical image segmentation. This review aims to discuss recent advances in FL for medical image segmentation with a focus on breast lesion and tumor segmentation.

One recent FL approach called FedZaCt, developed by T. Yang et al. \cite{FedZaCt2022}, combines Z-average aggregation with cross-teaching to improve image segmentation performance. The method was evaluated on the task of breast lesion segmentation and achieved superior performance compared to other FL methods. The Z-average aggregation reduces the impact of noisy updates from different devices, while cross-teaching encourages model diversity and enhances the ability of the models to generalize to new data.

Additionally, Z. Yang et al. \cite{yang2022robust} developed a Robust Split FL approach for U-shaped medical image networks that aims to improve the efficiency and robustness of FL for medical image analysis. The proposed approach includes a novel split learning method that partitions the model between the clients and the server to reduce the communication overhead. The proposed method was evaluated on a dataset of Brain Tumor Images and achieved superior performance compared to existing FL methods.

Another proposed approach by J. Wicaksana et al. \cite{FedMix:10004567} is a mixed supervised FL method for medical image segmentation that combines supervised and unsupervised learning to improve segmentation performance. The proposed method includes a supervised learning phase where the model is trained on labeled data and an unsupervised learning phase where the model is further refined on unlabeled data. The approach was evaluated on a dataset for USBCI and achieved promising results. The unsupervised learning phase helps to overcome the limitations of supervised learning by leveraging the large amounts of unlabeled data available in medical imaging. The use of unsupervised learning has been shown to improve the generalization of the model, making it more robust to unseen data.

For classification, Jiménez-Sánchez et al. \cite{JIMENEZSANCHEZ2023107318} introduced a new memory-aware curriculum FL approach for BC classification that improves the efficiency and performance of FL. The proposed method prioritizes the training of difficult samples by incorporating a curriculum learning approach. The proposed method was evaluated on a dataset for USBCI and achieved superior performance compared to existing FL methods.

In addition, M. M. Althobaiti et al. \cite{Althobaiti_Ashour_Alhindi_Althobaiti_Mansour_Gupta_Khanna_2022} also introduced a new Deep Transfer Learning-Based model for BC detection and classification using photoacoustic multimodal images. The proposed method aims to improve the accuracy and efficiency of BC diagnosis by training the model on multimodal images.

The use of such techniques can benefit medical professionals by enabling more accurate diagnoses and thus improving patient outcomes. FL has shown great potential in improving the accuracy and efficiency of medical image segmentation and classification. The proposed approaches aim to address the non-IID challenge in FL and improve the ability of the models to generalize to new data. These advances contribute to the standardization of medical image analysis and provide a benchmark for future studies in this field.

\section{Materials and Methods}

\subsection{Proposed Approach}

\begin{figure}\centering
  \includegraphics[width=1\textwidth,height=14pc]{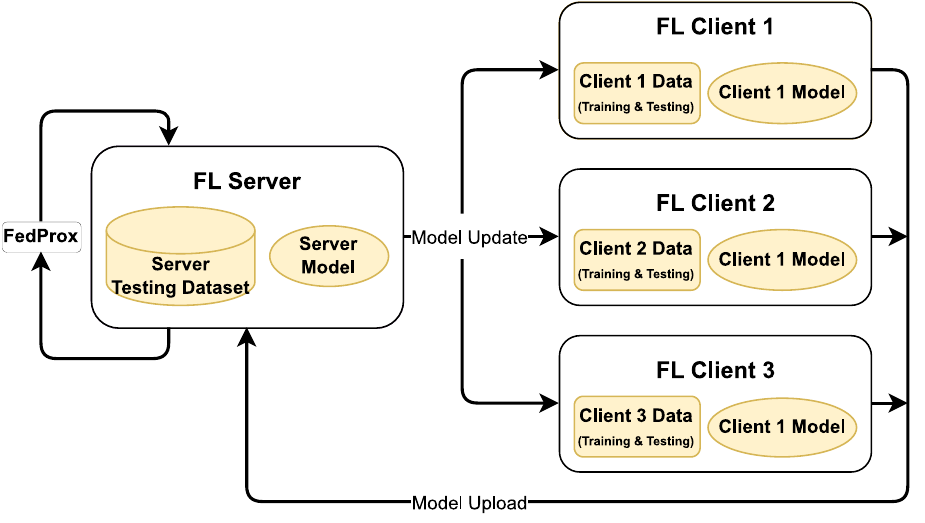}
  \caption{The Proposed Architecture of FL for BC Segmentation using Attention U-Net Model and FedProx}\label{fig:flmodel}
\end{figure}

In this study, we propose FL approach for BC segmentation, incorporating the FedProx method and an attention U-Net model. The architecture, as depicted in Figure \ref{fig:flmodel}, consists of a server side and a client side. On the server side, a global attention U-Net model is utilized, and the model's performance is evaluated using a testing dataset after each round. On the client side, three clients are set up, considering the limited availability of public USBCI datasets, each assigned with their respective data for training and testing, along with their attention U-Net models. Each client trains their models locally and shares the updated weights with the server. The FL Server then employs the FedProx algorithm to aggregate the updated weights. The newly aggregated weights are subsequently communicated back to the clients for further iterations. The proposed FL architecture aims to enable collaborative BC segmentation, ensuring data privacy, and improving model performance.

\subsection{Data Sources and Preparation}

This study utilized two datasets for our research. The first dataset, named "Dataset of breast ultrasound images," was annotated and published by Al-Dhabyani et al. \cite{Al-Dhabyani_Gomaa_Khaled_Fahmy_2020}. The second dataset, known as the "BUS B Dataset," was annotated and experimented with by Yap et al. \cite{Yap_Pons_Marti_Ganau_Sentis_Zwiggelaar_Davison_Marti_2018}. For simplicity, we will refer to the "Dataset of breast ultrasound images" as the BUS A dataset, while the BUS B dataset will retain its original name.

The BUS A dataset was collected in 2018 and provides a comprehensive collection of breast ultrasound images. The dataset includes scans from 600 female patients, totalling 780 scans, each accompanied by annotated masks. The images in this dataset are categorized into three groups based on the presence of normal, benign, or malignant features. It comprises 487 scans of benign cases, 210 scans of malignant cases, and 133 scans of normal cases. This dataset has been widely utilized in various studies for the development and evaluation of breast lesion classification and segmentation models. The BUS B dataset was collected in 2012 at the UDIAT Diagnostic Centre of the Parc Tauli Company in Sabadell, Spain, using a Siemens ACUSON Sequoia C512 system with a 17L5 HD linear array transducer \cite{Yap_Pons_Marti_Ganau_Sentis_Zwiggelaar_Davison_Marti_2018}. It consists of 163 scans of breast lesions along with their annotated masks. Among the scans, 110 belong to benign lesions, while 53 scans depict malignant tumors.

\begin{figure}\centering
  \includegraphics[width=1\textwidth,height=10pc]{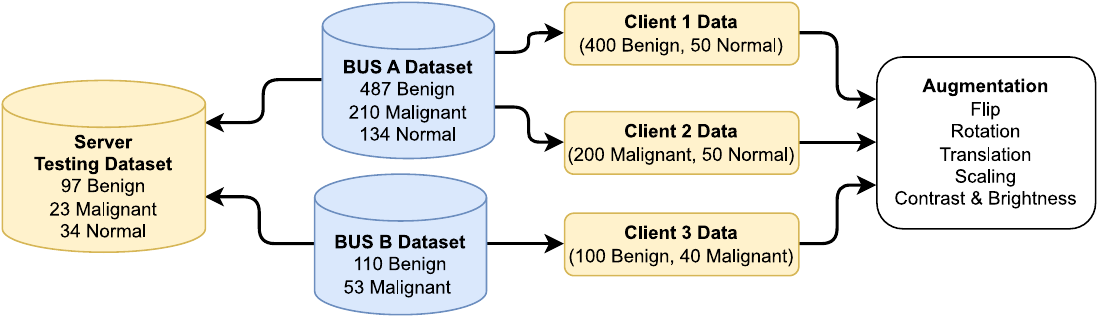}
  \caption{Distribution of BC data across the server and client sides}\label{fig:datadis}
\end{figure}

Figure \ref{fig:datadis} illustrates the division of data between the client side and the server side. To ensure a comprehensive evaluation of our approach, Each client contains distinct data with different features, representing various aspects of the problem. Client 1's data consists of 400 benign cases and 50 normal cases, while client 2's data includes 200 malignant cases and 50 normal cases, all sourced from the BUS A dataset. On the other hand, client 3 stands apart from the other clients and contains 110 benign cases and 53 malignant cases sourced from the BUS B dataset. In order to enhance the diversity and robustness of the training process, we applied various augmentation techniques to the clients' data. These techniques, including flip, rotation, translation, scaling, contrast adjustments, and brightness adjustments, were employed to increase the variability of the data samples. On the server side, a testing dataset is set up consisting of 97 benign cases, 23 malignant cases, and 34 normal cases, sourced from both datasets, enabling us to thoroughly evaluate the performance of the global model. This distribution of data across the clients and server ensures that the data is non-IID, allowing us to assess the effectiveness of our approach in handling diverse data and achieving accurate segmentation results of BC.

\subsection{FedProx}

The Fedprox method tackles the non-IID challenge in the data by partitioning it into non-overlapping subsets and distributing them to the clients. This ensures that each client has a representative sample of the data, which reduces the impact of distribution shift and improves the accuracy of the model. The proposed approach in this paper aims to enhance the accuracy and efficiency of the USBCI segmentation model by implementing Fedprox in a distributed system consisting of three clients and a server. The algorithm is designed to efficiently distribute the computational workload across multiple nodes, utilizing a distributed architecture to distribute resources effectively. The algorithm starts by initializing the model parameters on the server, and then the data is distributed to the clients. Each client performs local model updates based on its assigned data, and then sends the updated model parameters back to the server. \cite{li2020federated} The objective of the algorithm is to minimize the sum of the local loss functions of the clients with a proximal term and a regularizer term using the following formula:
\begin{equation}
    w^k = \arg\min_w \left(\sum_{j=1}^{n} \left(f_i(w) + \frac{\mu}{2} \|w - w_{i}^{k-1}\|_2^2\right)\right)
\end{equation}
Where $w^k$ is the model weights at iteration $k$, $f_i(w)$ is the local loss function of client $i$, $w_i^{k-1}$ is the model weights of client $i$ at iteration $k-1$, and $\mu$ is a hyperparameter that controls the strength of the proximal term. The proximal term enforces the model parameters to be close to the previous iteration, while the regularizer term promotes sparsity in the model. The resulting aggregated model parameters are sent back to the clients for further local updates. In this study, Fedprox was utilized for the USBCI segmentation task, with carefully chosen parameter values through experimental evaluation to optimize performance for the specific task, resulting in a learning rate of 0.01, a regularization parameter of 0.001, a proximal parameter of 0.01, and a hyperparameter value of $\mu = 0.1$. Fedprox has been shown to improve the performance of the distributed deep learning model on the USBCI dataset. Specifically, it addresses the inter-client variability of the dataset by encouraging the clients to learn from their local data while sharing the model updates.

\subsection{Attention U-Net}

In this study, The Attention U-Net is used in this study for segmenting BC tumors. The architecture of the Attention U-Net includes a contracting path with convolutional and max pooling layers followed by an expanding path with convolutional and up-sampling layers. The key difference from a standard U-Net is that it incorporates an attention mechanism in the skip connections. The attention mechanism enhances the network's ability to focus on relevant features while discarding irrelevant ones, resulting in improved segmentation accuracy.

\begin{figure}[!htb]
\centering
\includegraphics[width=1\textwidth]{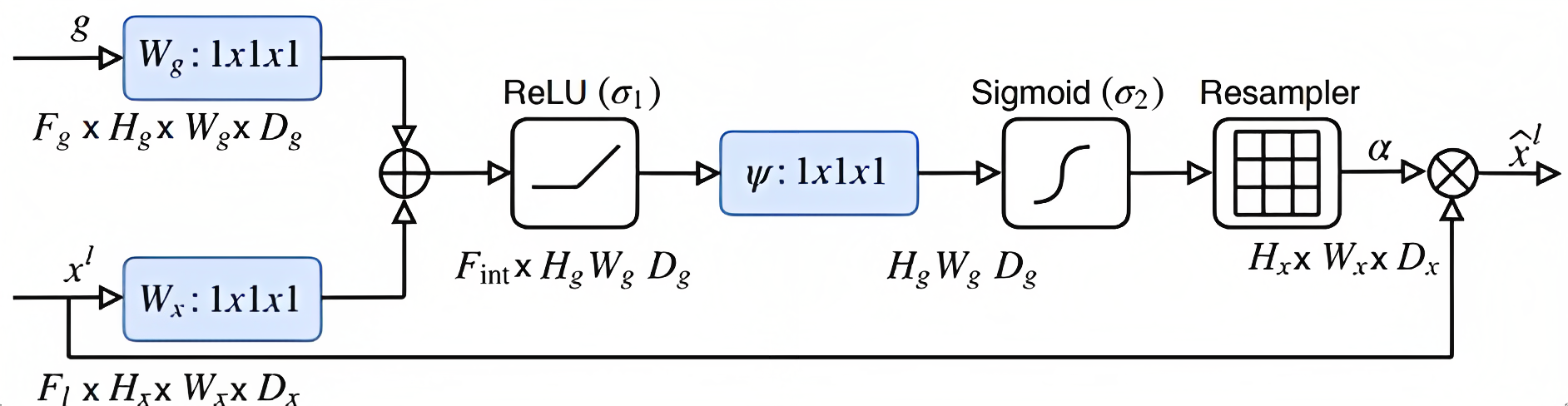}
\caption{The Attention Mechanism  \cite{oktay2018attention} }
\label{atten}
\end{figure}

The graphical model of the attention mechanism is presented in Figure \ref{atten}.\cite{oktay2018attention}, the attention mechanism works by taking two inputs, vectors X and G as shown in the following figure. Vector G is taken from the next lowest layer of the network and has smaller dimensions and better feature representation. Vector X goes through a strided convolution, and vector G goes through a 1x1 convolution. The two vectors are then summed element-wise, and the resulting vector goes through a ReLU activation layer and a 1x1 convolution that collapses the dimensions to 1x32x32. This vector then goes through a sigmoid layer, which scales the vector between the range (0-1), producing the attention coefficients (weights), where coefficients closer to 1 indicate more relevant features. The attention coefficients are unsampled to the original dimensions of vector X using trilinear interpolation. The attention coefficients are multiplied element-wise to the original vector X, scaling the vector according to relevance, and then passed along in the skip connection as normal. 

Attention U-Net model was implemented in the three clients with identical parameters. The training process utilized the dice loss function and the Adam optimizer with a learning rate of 0.0001 and a batch size of 16. The training was conducted over 10 epochs for each round, and evaluation was performed on a test set of 20\% of the local data for each client. The performance of the model was measured using the average dice coefficient and other relevant performance metrics.

\subsection{Performance Metrics}

To assess the performance of our model, we employed various performance metrics, including the confusion matrix. In this study, we calculated metrics, including Dice Loss, Intersection over Union (IoU), Sensitivity, Specificity, F1 Score, and Accuracy. These metrics were evaluated during both the training and validation phases of the system.

Dice Loss measures the dissimilarity between the predicted and true positive regions. It takes into account true positives (TP), true negatives (TN), false positives (FP), and false negatives (FN) \cite{V-Net:7785132}. Intersection over Union (IoU) quantifies the overlap between the predicted and true positive regions and is sometimes referred to as the Jaccard index \cite{Everingham_Van Gool_Williams_Winn_Zisserman_2009}.

Sensitivity, also known as recall or true positive rate, calculates the proportion of actual positives correctly identified by the model. It is computed using the following formula:
\begin{equation}
\text{Sensitivity} = \frac{\text{TP}}{\text{TP} + \text{FN}}
\end{equation}

Specificity measures the proportion of actual negatives correctly identified by the model and is calculated as:
\begin{equation}
\text{Specificity} = \frac{\text{TN}}{\text{TN} + \text{FP}}
\end{equation}

F1 Score is a balanced measure that combines precision and recall, providing an overall evaluation of the model's performance. It is computed using the following formula:
\begin{equation}
\text{F1 Score} = 2 \times \frac{\text{Precision} \times \text{Recall}}{\text{Precision} + \text{Recall}}
\end{equation}

Accuracy measures the proportion of correct predictions out of all predictions made by the model. It is calculated using the formula:
\begin{equation}
\text{Accuracy} = \frac{\text{TP} + \text{TN}}{\text{TP} + \text{TN} + \text{FP} + \text{FN}}
\end{equation}

These performance metrics provide insights into the effectiveness and accuracy of the model in capturing true positive regions and distinguishing between positives and negatives.

\section{Experimental Results}

In our proposed model, we conducted experiments to showcase the utilization of the FedProx algorithm and attention U-Net for the server model and for the client models, resulting in the development of a simple yet effective FL model. The training process consisted of 6 rounds, where each round involved training the client models over 10 epochs and subsequently aggregating the model using the FedProx algorithm on the server. To ensure unbiased outcomes, the performance of the global model in the server was evaluated using predefined metrics, while also calculating the metrics of the individual client models.
\begin{table}[!htbp]
  \centering
  \caption{Server (global model) results}
  \label{tab:performance}
  \begin{tabular}{lcccccc}
    \toprule
    Global & Dice Loss & IOU & Sensitivity & Specificity & F1 Score & Accuracy \\
    \midrule
    Round 1  &  0.895 & 0.0555 & 0.1575 & 0.8452 & 0.105 & 0.9204 \\
    Round 2  &  0.4364 & 0.3951 & 0.5214 & 0.9742 & 0.5636 & 0.9399 \\
    Round 3  &  0.3664 & 0.4639 & 0.5226 & 0.9897 & 0.6336 & 0.9525 \\
    Round 4  &  0.3282 & 0.5064 & 0.5842 & 0.9876 & 0.6718 & 0.9549 \\
    Round 5  &  0.3495 & 0.484 & 0.5518 & 0.9895 & 0.6505 & 0.9541 \\
    Round 6  &  0.2924 & 0.5494 & 0.6066 & 0.9919 & 0.7076 & 0.9607 \\
    \bottomrule
  \end{tabular}
\end{table}

Table \ref{tab:performance} presents a comprehensive summary of the server's (global model) performance across different training rounds. In the initial round (Round 1), the server model exhibited relatively high loss of 0.895 and low IoU of 0.0555, indicating a suboptimal level of segmentation accuracy. However, the model showed promise in specificity of 0.8452, suggesting its ability to correctly identify true negatives.
\begin{figure}[!htb]
\centering
\begin{subfigure}{0.45\linewidth}
\includegraphics[width=\linewidth]{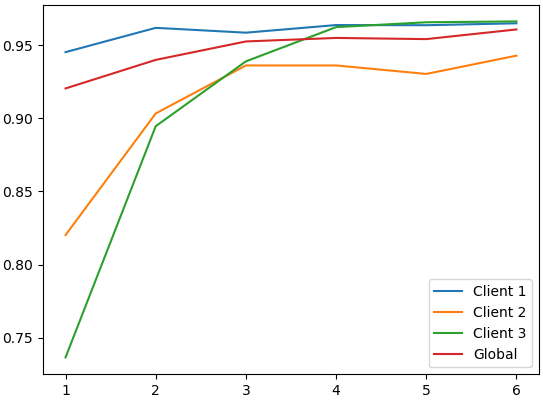}
\caption{Accuracy}
\label{fig:acc}
\end{subfigure}
\begin{subfigure}{0.45\linewidth}
\includegraphics[width=\linewidth]{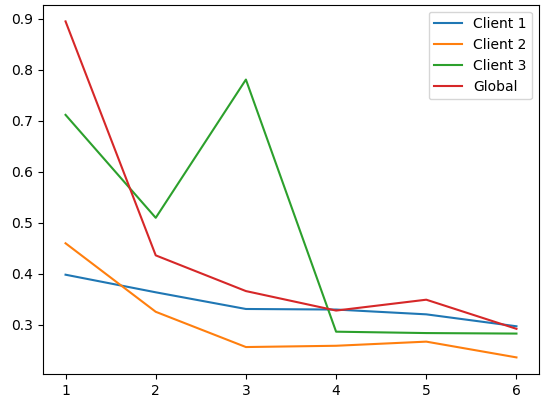}
\caption{Dice Loss}
\label{fig:loss}
\end{subfigure} \
\begin{subfigure}{0.45\linewidth}
\includegraphics[width=\linewidth]{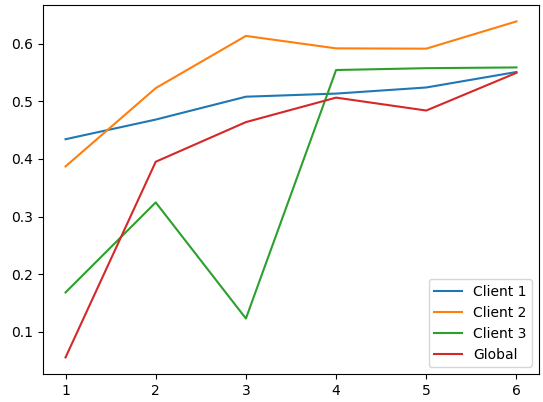}
\caption{IoU}
\label{fig:iou}
\end{subfigure}
\begin{subfigure}{0.45\linewidth}
\includegraphics[width=\linewidth]{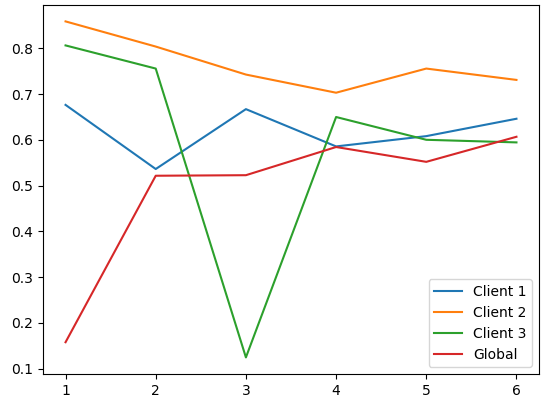}
\caption{Sensitivity}
\label{fig:sen}
\end{subfigure}
\caption{Performance Metrics of Server and Client Models}
\label{fig:grid}
\end{figure}

As training progressed to subsequent rounds, notable improvements in the server model's performance were observed. The loss values consistently decreased, indicating reduced overall error in the model. The IoU scores steadily increased, reflecting improved segmentation accuracy and better alignment with ground truth masks. Furthermore, the F1 scores, providing a balanced measure of precision and recall, showed a consistent upward trend, indicating overall performance enhancements.

By the final round (Round 6), the server model demonstrated significant advancements. It achieved a substantially lower loss of 0.2924, indicating a considerable reduction in error rate. The IoU score reached 0.5494, indicating a higher degree of overlap between predicted and ground truth masks. Additionally, the model exhibited excellent specificity of 0.9919, highlighting its proficiency in accurately identifying true negatives.

The performance of the models is visually represented in Figure \ref{fig:grid}, showcasing various metrics evaluated for the server model and three client models throughout the rounds of FL. The line graph \ref{fig:grid}(a) demonstrate the increasing accuracy of the FL global model as the rounds progress. In graph \ref{fig:grid}(b), the dice loss of the models exhibits a decreasing trend, indicating improved segmentation performance. graph \ref{fig:grid}(c) displays the IoU scores, showing an increasing trend and enhanced overlap between predicted and ground truth segmentation masks. Lastly, graph \ref{fig:grid}(d) presents the sensitivity of the models, with the FL global model showcasing a progressive increase, suggesting improved classification of positive instances.
\begin{figure}[!htb]
\centering
\includegraphics[width=\textwidth]{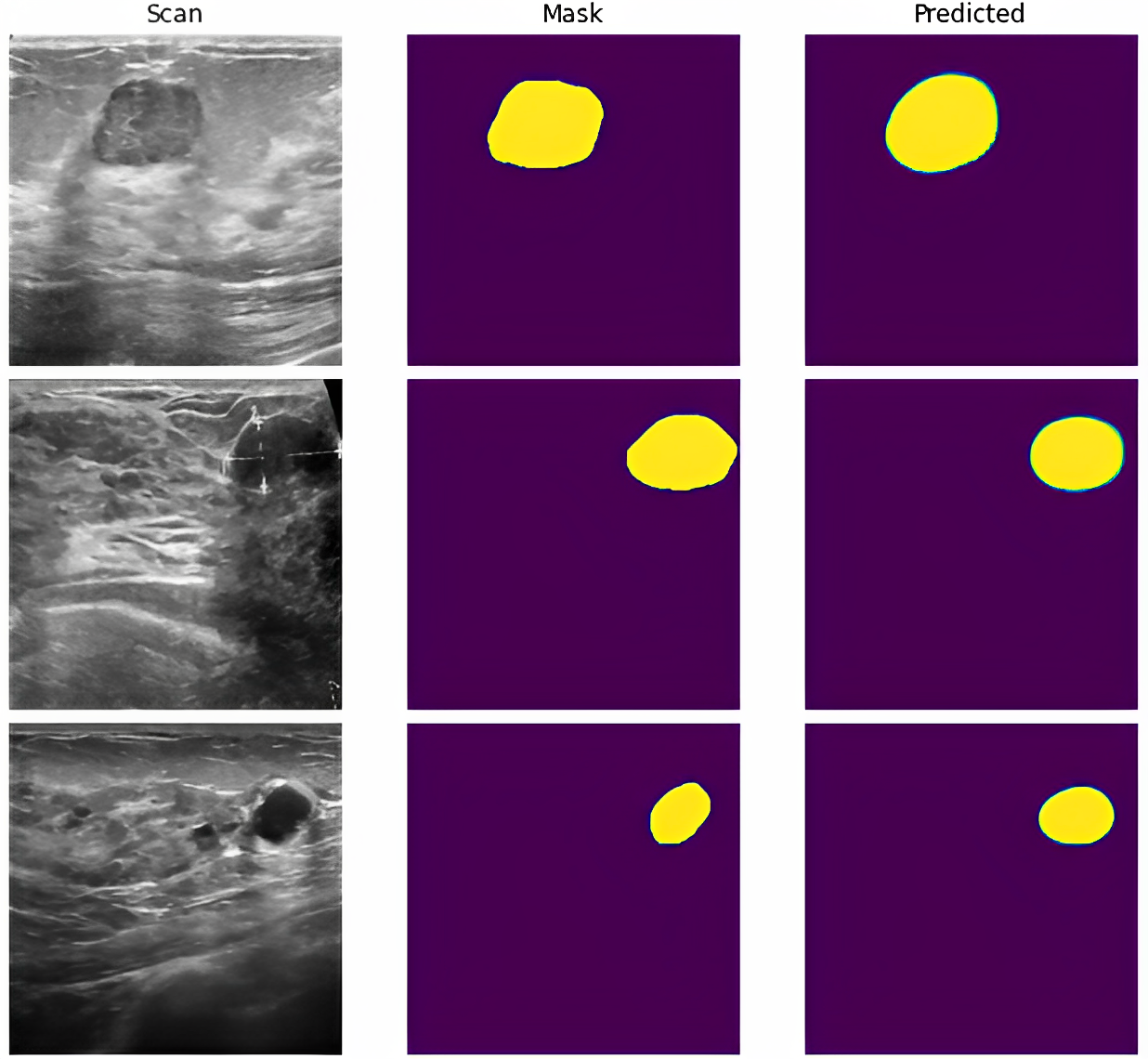}
\caption{Predicted outcome using Proposed Model} \label{viz}
\end{figure}

In Figure \ref{viz}, the visualization of the segmented global test scans obtained from our proposed model are displayed. The visualization highlight the high accuracy achieved by our model in segmenting the USBCI, with minimal noise observed in the predicted images. Upon comparing the predicted images with the ground truth images or mask images, it is evident that our model successfully captures the essential features of the lesions. This is demonstrated by the remarkable alignment between the boundaries of the predicted and ground truth images, emphasizing the model's ability to accurately identify and delineate BC lesions.

\section{Discussion}

The success of our proposed FedProx model in segmenting non-IID USBCI holds significant implications for medical image analysis, particularly in accurate BC diagnosis. By leveraging the Attention U-Net model, our method achieves a harmonious balance between local and global learning. The incorporation of attention mechanisms allows the model to focus on relevant features while disregarding irrelevant ones, resulting in improved segmentation accuracy. 
\begin{table}[htbp]
\centering
\caption{FedProx comparison with related studies}
\label{tab:comptab}
\begin{tabular}{lcccccc}
\hline
Study & Dice Loss & IOU & Sensitivity & Specificity & F1 Score & Accuracy \\
\hline
Camajori et al. \cite{9676574} & 0.78 & 0.78 & 0.65 & 0.88 & 0.98 & 0.78 \\
Yang et al. \cite{FedZaCt2022} & 0.84 & 0.77 & 0.87 & 0.98 & 0.82 & 0.96 \\
Wicaksana et al. \cite{FedMix:10004567} & 0.81 & 0.73 & 0.82 & 0.98 & 0.81 & 0.96 \\
Roth et al. \cite{RothHolger10.1007/978-3-030-60548-3_18} & 0.29 & 0.21 & 0.59 & 0.99 & 0.33 & 0.87 \\
Jiménez-Sánchez et al. \cite{JIMENEZSANCHEZ2023107318} & 0.59 & 0.44 & 0.72 & 0.94 & 0.64 & 0.90 \\
\textbf{This Study(FedProx)} & \textbf{0.29} & \textbf{0.55} & \textbf{0.64} & \textbf{0.99} & \textbf{0.71} & \textbf{0.96} \\
\hline
\end{tabular}
\end{table}
To validate the effectiveness of our proposed system, the performance of our proposed system was compared to other related studies, as shown in table \ref{tab:comptab}. It can be observed that our FL model outperforms or achieves comparable results to the related studies. It achieves a Dice Loss of 0.29, an IoU of 0.55, a sensitivity of 0.64, a specificity of 0.99, an F1 score of 0.71, and an accuracy of 0.96. These results demonstrate the effectiveness and potential of the FedProx model for accurate BC segmentation in medical imaging analysis, however, there is still room for improvement, as indicated by the varying performance metrics across the studies. Further research and refinement of the FedProx model can contribute to more accurate and efficient BC diagnosis.

\section{Conclusion}

Our study has demonstrated that the combination of USBCI and the FedProx algorithm can improve BC detection accuracy. The results indicate that the FedProx model is an excellent method medical image analysis across multiple devices while preserving privacy without sacrificing accuracy. The performance metrics of the proposed model supports this conclusion by determining that the model has an accuracy of 96\% in image segmentation, allowing a relatively small error to occur while setting the boundaries around the tumours. As such the boundaries will indicate the location of the tumour with high accuracy and regions of interest for the medical professionals to focus on. The proposed model demonstrated high specificity alongside the accuracy, indicating that it is capable of correctly identifying true negative cases. Overall, the proposed model proved to be a viable approach to medical image classification, allowing for better privacy preservation and improved model generalization, while still maintaining comparable performance to traditional model. These findings highlight the potential of using FL for medical imaging applications and the importance of exploring novel DL techniques to address challenges in the healthcare domain. Furthermore, we believe that our work contributes to the growing body of research on the application of FL in medical image analysis by improving BC detection.

%
%
%
%

\end{document}